\documentclass[10pt,twocolumn,letterpaper]{article}

\usepackage{graphicx}

\usepackage{wacv}
\usepackage{times}
\usepackage{epsfig}
\usepackage{graphicx}
\usepackage{amsmath}
\usepackage{amssymb}
\usepackage[section]{placeins}
\usepackage{multirow}
\wacvfinalcopy % *** Uncomment this line for the final submission

\ifwacvfinal
\def\assignedStartPage{1} % *** Enter the assigned starting page number (instead of 9876)
\fi

\ifwacvfinal
\usepackage[breaklinks=true,bookmarks=false]{hyperref}
\else
\usepackage[pagebackref=true,breaklinks=true,colorlinks,bookmarks=false]{hyperref}
\fi

\ifwacvfinal
\else
\fi

\begin{document}

%%%%%%%%% TITLE
\title{A Flow-Guided Mutual Attention Network for Video-Based \\Person Re-Identification}

\author{Madhu Kiran$^a$, Amran Bhuiyan$^{a}$, Louis-Antoine Blais-Morin$^b$, Mehrsan Javan$^{c}$,Ismail Ben Ayed$^{a}$, Eric Granger$^a$\\
$^a$ \small{\textit{Laboratoire d'imagerie, de vision et d'intelligence artificielle, \'Ecole de technologie sup\'erieure, Montreal, Canada}}\\
$^b$ \small{\textit{Genetec Inc., Montreal Canada}}\\
$^c$ \small{\textit{Sportlogiq Inc., Montreal Canada}}\\
{\tt\small mkiran@livia.etsmtl.ca, amran.apece@gmail.com}\\
{\tt\small lablaismorn@genetec.com, mehrsan@gmail.com} \\
{\tt\small  Ismail.BenAyed@etsmtl.ca, eric.granger@etsmtl.ca}
}

\maketitle
%\thispagestyle{empty}

%%%%%%%%% ABSTRACT
\begin{abstract}
\vspace{-0.3cm}
Person Re-Identification (ReID) is a challenging problem in many video analytics and surveillance applications, where a person's identity must be associated across a distributed non-overlapping network of cameras. Video-based person ReID has recently gained much interest because it allows capturing discriminant spatio-temporal information from video clips that is unavailable for image-based ReID. Despite recent advances, deep learning (DL) models for video ReID often fail to leverage this information to improve the robustness of feature representations. 
In this paper, the motion pattern of a person is explored as an additional cue for ReID. In particular, a flow-guided Mutual Attention network is proposed for fusion of image and optical flow sequences using any 2D-CNN backbone, allowing to encode temporal information along with spatial appearance information. 
%a new DL model -- comprised of a CNN backbone and a flow-guided attention module -- is proposed for ReID with spatio-temporal attention. 
%Given a query tracklet captured for an individual, the backbone CNN produces a deep feature embedding for pair-wise matching with that of reference tracklets, while the
%Our first proposed network -- called Gated Attention -- relies on optical flow to generate gated attention with video-based feature that embed spatially. Hence the proposed framework allows to activate a common set of salient features across multiple frames. 
Our Mutual Attention network relies on the joint spatial attention between image and optical flow features maps to activate a common set of salient features across them.
In addition to flow-guided attention, we introduce a method to aggregate features from longer input streams for better video sequence-level representation.
%In the paper, we propose using a new Mutual Attention Network  to capture visual appearance across both streams (video and optical flow). 
%Our method relies on the temporal stream to enhance both spatial information as well as temporal dynamics of video sequences. 
%
Our extensive experiments on three challenging video ReID datasets indicate that using the proposed Mutual Attention network allows to improve recognition accuracy considerably with respect to conventional gated-attention networks, and state-of-the-art methods for video-based person ReID. % while being able to process longer frame sequences.
%Additionally, our Mutual Attention network is able to process longer frame sequences with a wider range of appearance variations for highly accurate recognition.
\end{abstract}

%%%%%%%%% BODY TEXT
\section{Introduction}

%%%% APPLICATION
Person Re-Identification (ReID) refers to the problem of associating individuals over a set of non-overlapping camera views. It is one of the key object recognition tasks, that has recently drawn a significant attention due to its wide range of applications in monitoring and surveillance, e.g., multi-camera target tracking, pedestrian tracking in autonomous driving, access control in biometrics, search and retrieval in video surveillance, and human-computer interaction communities. Despite the recent progress with deep learning (DL) models, person re-identification remains a challenging task due to the non-rigid structure of the human body, the variability of capture conditions (e.g., illumination, blur), occlusions and background clutter.  

%%%% OVERVIEW OF LITERATURE
ReID systems can apply in image-based and video-based settings. State-of-the-art~\cite{ahmed2015improved,bhuiyan2020pose,bhuiyan2014person,bhuiyan2018exploiting,farenzena2010person,panda2017unsupervised,sun2018beyond,quan2019auto,tay2019aanet} approaches on image-based setting seek to associate still images of individuals over a set of non-overlapping cameras. In case of video based ReID where input video tracklets of an individual are matched against a gallery of tracklet representations, captured with different non-overlapping cameras. A tracklet corresponds to a sequence of bounding boxes that were captured over time for a same person in a camera viewpoint, and are obtained using a person detector and tracker. Compared to image data, video data provides richer source of information about persons' appearance along with motion information that notably capture persons' body layouts. Thus, video-based approaches allow to exploit spatio-temporal information (appearance and motion) for discriminative feature representation.
\begin{figure}[t]
 \centering
\includegraphics[width=1.0\linewidth]{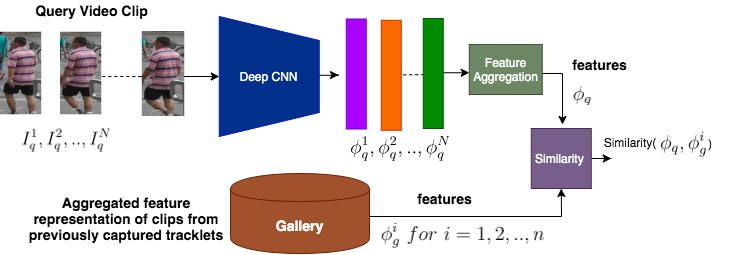}
   \caption{Block diagram of a generic DL model specialized for video-based person ReID. Each query video clip from a non-overlapping camera is input to a backbone CNN to produce a set of features embeddings, one per image. The features are then aggregated to produce a  single feature representation for the clip, which is then matched against clip representations stored in the gallery.}
  \label{fig:generic} 
   \vspace{-0.55cm}
\end{figure}

As illustrated in Figure~\ref{fig:generic}, state-of-the-art approaches for video-based person ReID typically learn global features in an end-to-end fashion, through various temporal feature aggregation techniques~\cite{revisiting,temporaliccv,occfree,LSTM,coseg}. From this figure, the query input to the feature extractor is a video clip(set of bounding boxes extracted from a tracklet) of $N$ frames long. A single feature vector is extracted by aggregating features from each frame in the query video clip. This is then compared with a gallery containing $n$ identities which is  set of aggregated feature representations of clips from previously-captured tacklets. 

 Given a video clip (fixed size set of bounding boxes extracted from a tracklet), the feature extractor (CNN backbone) produces image-level features, while the feature aggregator generates a single feature representation at the clip level, using either average pooling, weighted addition, max pooling, recurrent NNs, etc.~\cite{revisiting} in the temporal domain. Although these aggregation approaches enable to incorporate diverse tracklet information for matching, and can achieve a higher level of accuracy that image-based approaches, they often fail to efficiently capture temporal information which could propagate as salient features throughout the video sequence. 
Additionally, the performance of state-of-the-art methods decline as the length of video clips grows beyond 4 or 6 frames~\cite{revisiting,coseg}. 

\begin{figure}[t!]
 \centering
\includegraphics[width=0.6\linewidth]{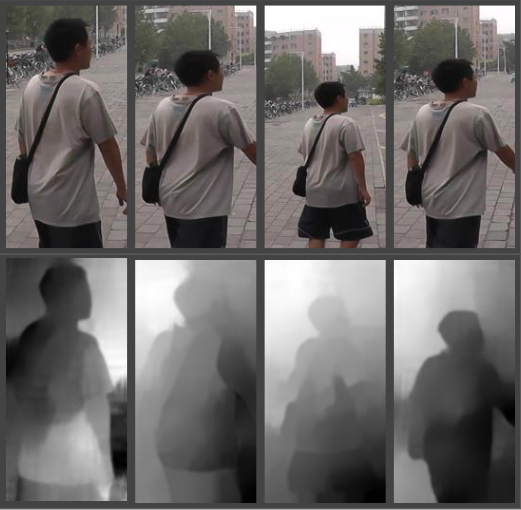}
   \caption{Example of a sequence of bounding boxes images from the MARS dataset (top row), and its corresponding dense optical flow map (bottom row). The common saliency in the sequence can be observed from the optical flow map.}
  \label{fig:exemplar}
   \vspace{-0.55cm}
\end{figure}

 Optical flow stream  has been previously used as additional stream of input captures the motion dynamics of a walking person in a video stream. At the same time, as shown in Fig.~\ref{fig:exemplar}, visual appearance of optical flow for a walking or moving person is very close to the silhouette of the person often suppressing the background static objects. This has a potential to be able to be used as a mask on appearance stream to highlight common saliency between frames. 
 The potential silhouette produce by optical flow together with it highlighting common saliency across frames can therefore be a good source of spatio-temporal attention. 
  
 Fig.~\ref{fig:exemplar} shows that the flow features are coarse representation of semantic information of moving objects. Unlike action recognition which depends heavily on motion features, ReID is more dependent on appearance features. Hence, there is a scope to combine the strengths of optical flow and appearance (video stream) features for ReID. Previous attempts to include optical flow information into ReID systems~\cite{optical1,optical2,optical3} focused on integrating this information as an additional input in the network with some kind of integration into the main features later. This is not effective because optical flow only represents coarse semantic features of moving objects (different from the image stream), and not image-like appearance information. Moreover the model in~\cite{optical1} is related to a two-stream network proposed in~\cite{simonyan2014two} that incorporates motion and appearance feature for action recognition. Two-stream networks that are effective for action recognition are less effective for ReID~\cite{optical1}. 
 
 Given the aforementioned justification, in this work the correlation between the visual appearances across motion and appearance stream along with their individual contribution to motion dynamics are considered. In order to capture long term spatial information and temporal dynamics in a video clip, a method to aggregate features from longer sequence effectively is presented. This has not been explored in the literature for video person-ReID, thereby undermining the global saliency in the feature representation by using optical flow for both appearance and motion information. 

%allowing to encode temporal information along with spatial appearance information for ReID. 
In this paper, DL model for flow-guided attention is introduced for video-based person ReID to enable joint spatial attention  between input temporal(optical flow) and image stream (video sequence). The proposed Mutual Attention network enables to jointly learn a feature embedding that incorporates relevant spatial information from human appearance, along with their motion information, from both appearance stream (images) and motion stream (optical flow).
%Figure~\ref{fig:flow} presents the overall architecture The proposed Gated Attention model. 
%
%--Given a query tracklet captured for an individual, the backbone CNN produces a deep feature embedding for pair-wise matching with that of reference tracklets, while the.
%Our first proposed network -- called Gated Attention -- relies on optical flow to attend spatially to image feature. We propose to integrate optical flow-guided attention map to generate a video based feature embedding  with spatial attention. A feature aggregation module is also introduced to generate video descriptor propelled by flow guided temporal attention. Hence the proposed framework allows to activate a common set of salient features across multiple frames. 
The Mutual Attention network includes both optical flow stream and image stream for ReID and leverages the mutual appearance and motion information. 

We also propose a feature aggregation method to capture long-range temporal relationship by being able to aggregate information from longer video tracklets or sequences.
%In contrast, our second network -- called Mutual Attention -- relies on the the joint attention between image and optical flow features. This enables spatial attention between both sources of features, across motion and appearance cues. 
Unlike prior work in literature where feature aggregation is achieved by pooling or temporal attention from image feature, the proposed Mutual Attention network relies on a weighted feature addition method over images in a sequence to produce a single feature descriptor using both optical flow and image feature information. During feature aggregation, a reference frame from each tracklet is selected based on maximum activation from both the streams, and weights are assigned for individual features using image and flow feature information. Attention is enabled from optical flow in both spatial and temporal domain to extract discriminant features for ReID. 

Performance of the proposed flow-guided attention network is evaluated and compared on the challenging MARS, Duke-MTMC datasets and ILIDS-Vid dataset for video-based person ReID. Experimental results show that both improve accuracy and can outperform the state-of-the-art approaches. Results also indicate the capability for higher accurate predictions by using longer video clips  to capture multiple appearance variations.

%%%%%%%%%%%%%%%%%%%%%%%%%%%%%
\section{Related Work}

The section provides background on DL models for spatio-temporal recognition, optical flow, and attention mechanisms as they relate to person ReID.

%%Amran
\noindent \textbf{a) Image-Based Person-ReID:} 
The idea of using CNNs for ReID stems from Siamese Network~\cite{bromley1994signature}, which involves two sub-networks with shared weights, and is suitable for finding the pair-wise similarity between query and reference images. It has first been used in~\cite{yi2014deep} that employs three Siamese sub-networks for deep feature learning. Since then many authors focus on designing various DL architectures to learn discriminative feature embedding. Most of these deep-architecture based 
ReID~\cite{ahmed2015improved,cheng2016person,varior2016gated,chen2017beyond,chen2017beyond,liu2017end} approaches introduce an end-to-end ReID framework, where both feature embedding and metric learning 
have been investigated as a joint learning problem. In~\cite{ahmed2015improved,varior2016gated}, a new layer is proposed to capture the local relationship between two images, which helps modeling pose and 
viewpoint variations in cross-view pedestrian images. Recent ReID approaches~\cite{su2017pose,suh2018part,zheng2017pose,zhao2017spindle,qian2018pose,zhao2017deeply,saquib2018pose,sun2018beyond,bhuiyan2020pose} rely on incorporating contextual information into the base deep ReID model, where local and global feature representations are combined to improve accuracy. A few attention-based approaches for deep re-ID ~\cite{li2017learning,zhao2017deeply,su2017pose} address misalignment challenges by incorporating a regional attention sub-network into a base re-ID model. A thorough review of state-of-the-art on architecture-based approaches underscores the importance of considering local representations, e.g., by dividing the image into soft stripes~\cite{sun2018beyond} or by pose-based part  representation~\cite{su2017pose,suh2018part,zheng2017pose,zhao2017spindle,qian2018pose,zhao2017deeply,saquib2018pose}. Although these methods have have achieve considerable performance improvements, they fail to incorporate temporal information due to their image-based setting.  

%%Madhu
\noindent \textbf{b) Video-Based Person-ReID:} 
 Video ReID has recently attracted some interest since temporal information allows dealing with ambiguities such as occlusion and background noise~\cite{revisiting,temporaliccv,coseg,occfree,LSTM}. An important problem in video-based ReID is the task of aggregating the image level features to obtain one single composite feature or descriptor for a video sequence. \cite{revisiting} have approached this problem by frame level feature extraction and temporal fusion by using recurrent NNs (RNNs), average pooling, and temporal attention (based on image features). Average Pooling in temporal dimension can be  viewed as summing the features of the sequence by giving equal normalised weights to them. Average pooling of image instance features from a given sequence have proved to be useful in most of the cases, even compared to other DL model based on RNNs or 3D-CNN~\cite{revisiting}. 3DCNN has been experimented in~\cite{revisiting,3dcnn} but have not been very effective in summarising video sequence for reID. But there could be certain case of individual image in a sequence such that they either have higher noise content or the appearance in the image does not contribute much to an individual's identity, then these become the debatable cases for Average Pooling. 

\noindent \textbf{c) Attention Mechanisms:}
Attention can be interpreted as a means of biasing the allocation of available computational resources towards the most informative components of a signal~\cite{sen}.   A mask guided attention mechanism has been proposed in~\cite{maskguided}, where a binary body mask is used in conjunction with the corresponding person image to reduce background clutter. Somewhat similar to~\cite{maskguided}, co-segmentation networks have achieved significant improvements in ReID accuracy over the baseline by connecting a new COSAM module between different layers of a deep feature extraction network~\cite{coseg}. Co-Segmentation allows extracting common saliency between images, and using this information for both spatial- and channel-wise attention. Other related work for attention in video ReID , ~\cite{STAL} attention is employed in both temporal and and spatial domain. Video stream has been taken advantage of by ~\cite{rquen} by extracting complementary region based feature by from different frames to obtain informative features as a whole.

\noindent \textbf{d) Optical Flow as Temporal Stream:}
  It often serves as a good approximation of the true physical motion projected onto the image plane~\cite{optical3}. Optical flow has been employed for temporal information fusion in~\cite{optical1,optical2}, in a two stream Siamese Network with a weighted cost function to combine the information from both the streams. It uses a CNN that accepts both optical flow and color channels as input, and a recurrent layer to exploit temporal relations. Its important to note that prior to ~\cite{optical1}, ~\cite{simonyan2014two} have used two stream networks but for action recognition. Two stream networks on their own are useful in action recognition as impact of motion cues in action recognition are higher in action recognition than that of ReID~\cite{optical1}. Therefore there is a necessity to use optical flow in a way that it can be leveraged for appearance related task. However, traditional two-stream networks are unable to exploit a critical component in re-id i.e appearance across both optical flow and image stream together. Similar to our motivation for using optical flow for appearance along with motion has been discussed in ~\cite{ts_lstm}. Similar to to our work, motivation for considering long term temporal relationship has been discussed in ~\cite{st_fusion}.

It can be summarised from the above that, as discussed in \cite{coseg}, \cite{dual_attention},\cite{maskguided} and \cite{dual_attention}, various saliency feature enhancement methods have been attempted, and in most cases, they help improve the overall performance. Optical flow typically encodes motion information in contrast to appearance information, and hence there is scope to explore enhancing appearance information from motion and vice-versa. Also, based on~\cite{st_fusion} and~\cite{ts_lstm}, highlight the advantages of long term temporal information and appearance across both optical flow and image stream together.

%%%%%%%%%%%%%%%%%%%%%%%%%%%%%%%%%%
\section{Proposed Mutual Attention Network}

Given an input video clip (set of consecutive bounding boxes extracted from a tracklet) represented by  $\mathbf{I_c^1,I_c^2,...,I_c^n}$ and corresponding optical flow estimations $\mathbf{F_c^1,F_c^2,...,F_c^n}$ where $c$ indicates the Identity of the video clip of length $n$, our objective is to extract a discriminative feature vector $\boldsymbol{\phi_c}$ for ReID. 

A new model for flow guided attention --  Mutual Attention network is proposed. It learns spatial-temporal attention from optical flow thereby focusing on common salient features of a given person during its motion across consecutive frames of a given video clip.
Although a two stream Siamese network has been proposed in~\cite{optical1}, they have included optical flow as an input for re-identification, and do not exploit the full potential of this information. Hence as discussed earlier, the motivation behind this work is to take advantage of visual appearance of both spatial and temporal stream i.e image and optical flow stream by producing a correlation map between them in the feature space. This correlation map is used as attention on both the input streams. The temporal information in both the streams is enhanced by enabling the use of longer video and optical flow clips with our proposed feature aggregation method.

Therefore our Mutual Attention network includes both optical flow and image streams which attend each other to obtain Mutual Attention and also to combine the features to yield a single feature representation per input pair of image and optical flow clip. 

%%%%%%%%%%%
\begin{figure}[t]
 \centering
  \includegraphics[width=1.0\linewidth]{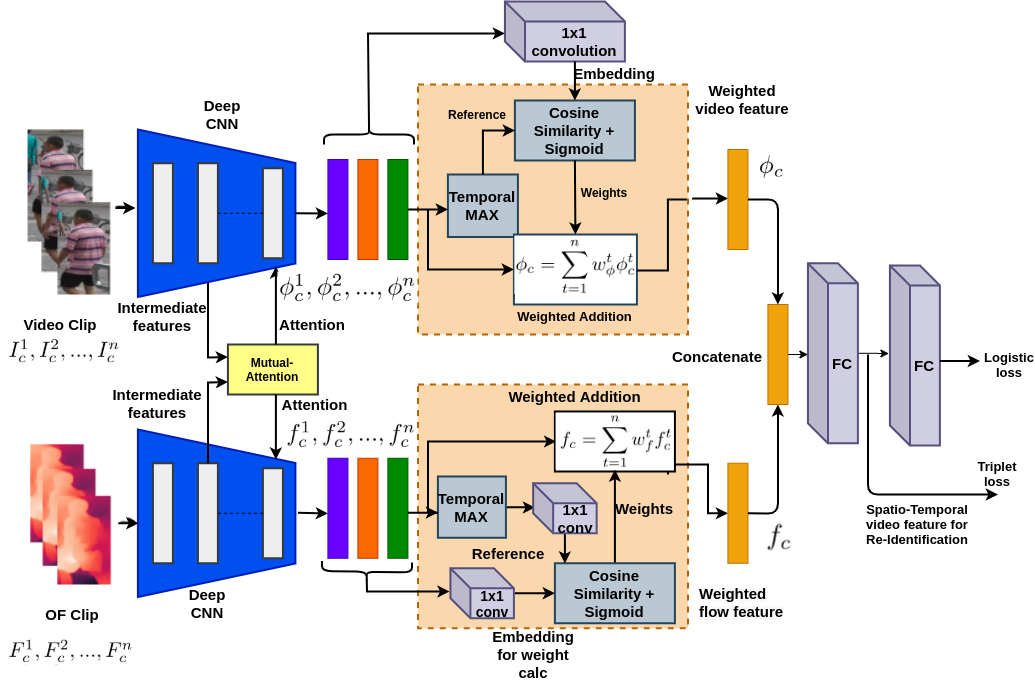}
   \caption{Our proposed Mutual Attention network. The input to the system is a video clip and corresponding flow maps. The features extracted from each other apply Mutual Attention to each other. The network outputs a concatenated feature from both optical flow and image stream.  }
  \label{fig:main}
   \vspace{-0.55cm}
\end{figure}

This is illustrated in Fig.~\ref{fig:main}, where the network accepts two streams of input (optical flow and image sequences). At the last layer of the network, the features from the two streams are concatenated after feature aggregation in the temporal domain. While the image stream helps in ReID by focusing on the appearance of the person, optical flow stream helps by capturing motion pattern of a given person. We propose to achieve feature aggregation to produce a feature vector by weighted addition. Our proposed method handles generation of weights that indicate the importance of individual image feature in producing a single video feature leveraging upon mutual attention.

In contrast to the previous method for flow guided attention mention above, we propose to produce cross-stream attention or Mutual attention between the optical flow stream and image stream  to boost areas in the feature space that have high activation across both the streams.

Given a video clip and corresponding flow maps, we extract the features $\boldsymbol{\phi_c}$ and $\mathbf{f_c}$ from the deep CNNs respectively. The expected output is a concatenated feature vector of both optical flow and image features to be used for ReID. Both the CNNs share common architecture but do not share the parameters. Let $l$ be the intermediate layer of the $k$-layer deep CNN and let appearance CNN be represented by $H_{\mbox{app}}$ and optical flow stream CNN by $H_{\mbox{flow}}$ with a  total number of $k$ layers. With $t={1,2,3...n}$ we have:
\begin{equation}
\label{eqn:layer}
        \boldsymbol{\phi_c^t} = H_{\mbox{app}}(\mathbf{I_c^t})~~,~~ 
        \mathbf{f_c^t} = H_{\mbox{flow}}(\mathbf{F_c^t})
\end{equation}
If features from layer $l$ are expressed as $\mathbf{\phi^l}$, then 
\begin{equation}
\label{eqn:intermediate}
\boldsymbol{\phi_c^{l,t}} = H_{\mbox{app,l}}(\mathbf{I_c^t})~~,~~ 
\mathbf{f_c^{l,t}} = H_{{\mbox{flow,l}}}(\mathbf{F_c^t}) 
\end{equation}
Both the features at layer $l$ are of dimensions, $N \times C\times I \times J$, representing sequence length,channels, width, height respectively. The features are then passed through $1x1$ convolution with $Relu$ activation to produce a map of size $N \times 1 \times I \times J$  each. The correlation between the features is given by, 
\begin{equation}
\label{eqn:correl}
       \boldsymbol{\rho} = \zeta_{\mbox{app}}(\boldsymbol{\phi_c^{l,t}})\odot\zeta_{\mbox{flow}}(\mathbf{f_c^{l,t}})
\end{equation}
In the Eq.~\ref{eqn:correl}, $\zeta_{\mbox{app}}$ and  $\zeta_{\mbox{flow}}$ are the embeddings with $\mathrm{1}\mathrm{x}\mathrm{1}$ convolution with $Relu$ discussed above. $\boldsymbol{\rho}$ when activated by a sigmoid function forms the mutual attention map $M_c^t$ between both the streams of input.

\begin{equation}
\label{eqn:activation}
       \mathbf{M_c^t} = \frac{\mathrm{1} }{\mathrm{1} + e^- \boldsymbol{\rho} }
\end{equation}
Finally, mutual attention is applied to the intermediate features $\boldsymbol{\phi_c^{l,t}}$ and $\mathbf{f_c^{l,t}}$ at the intermediate layer (by an element-wise multiplication of attention map with feature maps) to obtain mutually attended appearance features $\boldsymbol{\Psi_{app}}$ and $\boldsymbol{\Psi_{flow}}$ to continue feature extraction continues in the remaining layers of the deep CNN to obtain final output features  $\boldsymbol{\phi_c^t}$ and $\mathbf{f_c^t}$ for image and flow stream, respectively:
\begin{equation}
\label{eqn:attention}
    {\boldsymbol{\Psi_{app}}_c^t} =\boldsymbol{\phi_c^{l,t}} \odot  \mathbf{M_c^t}~~,~~ 
    {\boldsymbol{\Psi_{flow}}_c^t} =\mathbf{f_c^{l,t}} \odot  \mathbf{M_c^t}
\end{equation}

\paragraph{Weighted Feature Addition.}
We propose a method to aggregate  image level features to obtain a single feature vector for a given video sequence particularly enabling to use longer video sequences. The appearance features and optical flow features are then concatenated to for ReID during inference, and to learn a classifier during training. 

The output from image and optical flow stream CNNs generate $\boldsymbol{\phi_c}$ for a sequence $c$ from instances $\boldsymbol{\phi_c^1,\phi_c^2,...,\phi_c^n}$  and  $\mathbf{f_c}$ from  $\mathbf{f_c^1,f_c^2,...,f_c^n}$. The first task is to identify salient feature from a given sequence of features. In our case, a salient feature can be defined as the one that has maximum activation in both image and flow stream. Since the features have been attended by mutual attention, given a sequence, a max operation in the temporal domain for each of the sequence will identify the salient feature among the sequence. We hereafter will refer to this salient feature as reference frame denoted by $\boldsymbol{\phi_c^{max}}$ and $\mathbf{f_c^{max}}$. In the next step an adaptive weight is generated for each of the features in the sequence based on how close each feature is with the reference feature. This is achieved by applying a cosine similarity between the reference feature and rest of the features int he sequence. The cosine similarity function is not applied directly on the features $\boldsymbol{\phi_c^n}$ and $\mathbf{f_c^n}$. Instead a tiny embedding $\boldsymbol{\epsilon(.)}$ is applied on the  $\boldsymbol{\phi_c^n$,  $f_c^n}$ and reference feature $\boldsymbol{\phi_c^{max}}$, $\mathbf{f_c^{max}}$  to obtain embeddings $\boldsymbol{\phi_{\epsilon}^n}$, $\mathbf{f_{\epsilon}^n}$, $\boldsymbol{\phi_\epsilon^{max}}$ and $\mathbf{f_\epsilon^{max}}$:
\begin{equation}
\label{eqn:Appweight}
  w_{app}^{n}=\exp \left(\frac{\boldsymbol{\phi_{\epsilon}^n}\cdot \boldsymbol{\phi_\epsilon^{max}}}{\left||\boldsymbol{\phi_{\epsilon}^n}\right|\left|\boldsymbol{\phi_\epsilon^{max}}\right| \mid}\right)
\end{equation}
\begin{equation}
\label{eqn:flowW}
    w_{flow}^{n}=\exp \left(\frac{\mathbf{f_{\epsilon}^n}\cdot \mathbf{f_\epsilon^{max}}}{\left||\mathbf{f_{\epsilon}^n}\right|\left|\mathbf{f_\epsilon^{max}}\right| \mid}\right)
\end{equation}

\begin{equation}
\label{eqn:mil1}
\boldsymbol{\phi_c}=\sum_{t=1}^{n} w_{app^{n}} \boldsymbol{\phi_c^n}
\end{equation}

From Eq.~\ref{eqn:Appweight} and Eq.~\ref{eqn:flowW}, feature aggregation weights are calculated for both image features and flow features respectively. The features are aggregated as per weighted aggregation Eq.~\ref{eqn:mil1} to form outputs $\boldsymbol{\phi_c}$ and when used with $ w_{flow}$ to obtain $\mathbf{f_c}$ which are aggregated video features for image and flow respectively. These two features are concatenated to form $\boldsymbol{\phi_{cat}}$ which is passed though a fully connected layer of size same as $\boldsymbol{\phi_c}$ to produce the final feature for classification or re-identification.
The network is trained on logistic loss and Triplet loss similar to the method used in ~\cite{coseg}. During testing the fully connected layer is removed and the remainder of the network is used for feature extraction purpose and to match against those in gallery. 

%\subsection{Overall Architecture}
%A variety of networks have been previously proposed to extract invariant features for ReID in the literature. We follow the overall system architecture as ~\cite{revisiting} (Baseline) and ~\cite{coseg} (COSAM). They achieved state-of-the-art results on several ReID datasets. They have used ResNet50 for feature extraction. We propose to use ResNet50 as our base network to learn features invariant to cluttered background by attending with saliency map obtained from optical flow estimations. 
%We experiment at different layers of the ResNet50 to select the ideal location in the network to generate maximum attention. To extract video level feature from instance level features, we compare our proposed feature aggregation (FA) with that of temporal Average Pooling (AP) and Temporal Attention (TA) based method as illustrated in ~\cite{revisiting} and \cite{coseg}. For flow feature extraction for gated attention, we use a three layer CNN with 3x3 filters. Max Pooling has been used between layers along with Relu activation function. For our Mutual Attention method we use a full ResNet similar to the image feature extractor used in both the Methods.

%%%%%
\section{Experimental Methodology}

In this section the experimental procedure is described to evaluate and compare the performance  of proposed and reference feature aggregation methods. The implementation details describe the datasets, feature extraction models, parameters for flow feature estimation, optical flow estimation method, training and evaluation techniques.
We first aim to compare the benefits of Mutual attention with two stream network against one stream network simple Gated Attention with optical flow. We also compare our feature aggregation model with other methods in Video ReID literature including averaging, temporal attention and Recurrent networks.

\vspace{-2mm}
\paragraph{Datasets.} Experiment are performed on 2 challenging and widely-used datasets for video-based person reID. MARS~\cite{mars} dataset is one of the largest datasets for video Person-ReID.  Another dataset commonly used in literature for evaluating video person ReID is Duke-MTMC~\cite{duke1},~\cite{duke2} dataset  containing 702 identities and more than 2000 sequences for testing and training each. We also evaluate on ILIDS-VID~\cite{ilids} dataset, which has a total of 300 identities with videos across two cameras. One interesting point to be noted with ILIDS datasets is that the tracklets have been generated by hand annotation unlike detector based annotation in MARS dataset. This makes the bounding boxes well aligned in ILIDS-VID dataset enabling the optical flow estimation to be less noisy.

\vspace{-3mm}
\paragraph{Deep Feature Extraction.}  We follow the overall system architecture in~\cite{revisiting} (Baseline) and ~\cite{coseg} (COSAM). They achieved state-of-the-art results on several ReID datasets. They have used ResNet50 for feature extraction. We propose to use ResNet50 and SE-ResNet 50 as our base networks to learn features invariant to cluttered background by attending with saliency map obtained from optical flow estimations. We have shown results with each of the networks as back end separately. The networks have been pre-trained on imagenet~\cite{imagenet_cvpr09} dataset. We experiment at different layers of the ResNet50 to select the ideal location in the network to generate maximum attention with optical flow. To extract video level feature from instance level features, we compare our proposed weighted addition method with that of temporal Average Pooling (AP) and Temporal Attention (TA) based method as illustrated in ~\cite{revisiting} and \cite{coseg}. The Shallow CNN in our experimental setup for gated attention is based on AlexNet and the sub-Network for weighted addition is a two layer MLP of size 2048 nodes in each layer.

\vspace{-3mm}
\paragraph{Optical Flow Estimation.} To estimate optical flow maps for a given sequence  LiteFlowNet~\cite{litefloenet} model has been chosen as they are computationally efficient compared to other deep models along with obtaining state of the art performance. We have used the official implementation from the authors to produce flow maps for both MARS and Duke-MTMC dataset. Hence for a given sequence, we input pairs of image $I^{t-1}, I^t$ as input to the LiteFlowNet model to produce flow map $F^{t-1}$. 

\vspace{-3mm}
\paragraph{Evaluation Measures.} During the training phase we learn the ReID task by training a classifier with identity labels from the single feature extracted for a sequence. The feature tractor produces a $2048\times1$ size feature vector per sequence. This is the input to train the ReID classifier. During the testing phase, we use the 2048 dimensional feature to measure distance between the test sequence and the sequence from the gallery. We use the Cumulative Matching Characteristic(CMC) and Mean Average Precision(mAP) to evaluate the performance. CMC represents the matching characteristics of the first $n$ query results.

\vspace{-3mm}
\paragraph{Settings.} The network feature extractors have been pre-trained on ImagNet dataset. We follow common data augmentation such as random flips and random crops during training. We use ADAM optimizer to train our model with a batch size of 32. We use a sequence length of 4 to train our model. The flow guided attention has been applied at layer 4 of the ResNet50 and an empirical study of attention at different layers has been presented in the next section. Hence the training setting have mostly been kept similar to our baselines~\cite{revisiting}. In order to explore the advantage of attention with optical flow, we propose a separate experimental setup with simple gated attention mechanism i.e using optical flow to attend to image stream alone. This is a one stream approach where only image features are used for classification while optical flow is just used as an attention mechanism as described below.

\begin{figure}[t]
 \centering
\includegraphics[width=1.0\linewidth]{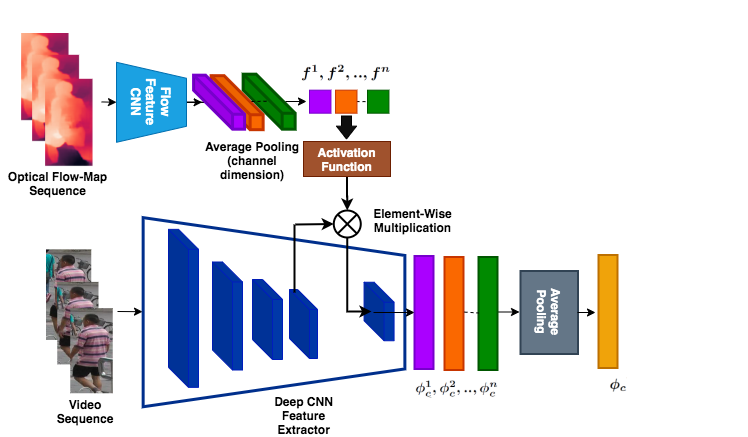}
   \caption{Our experimental setup for our baseline  gated attention network. The system inputs a sequence of bounding boxed images and the corresponding optical flow maps from a given video clip. The features extracted from optical flow are pooled in channel dimension, and multiplied with intermediate layers of deep feature extraction after activation to obtain attended features for ReID. The network outputs a feature vector  $\phi_c$ per video clip.}
  \label{fig:flow} 
   \vspace{-0.55cm}
\end{figure}

Given input images, {$\mathbf{I_c^1,I_c^2,...,I_c^n}$} and flow maps {$\mathbf{F_c^1,F_c^2,...,F_c^n}$} for images of a sequence, we extract the features $\boldsymbol{\phi_c}$ = ($\boldsymbol{\phi_c^1,\phi_c^2,...,\phi_c^n}$) and $\mathbf{f_c}$ = ($\mathbf{f_c^1,f_c^2,...,f_c^n}$) from the deep CNN and the Flow CNN respectively. A shallow CNN (Flow CNN) has been used to extract features from optical flow maps. We rely on shallow CNN to retain spatial coherence in the flow features (see  Fig.~\ref{fig:flow}). 

Let $l$ be the intermediate layer of the $k$ layer deep CNN and let the deep CNN be represented by $H$ with a total number of $k$ layers. Let $S$ denote the shallow flow feature extractor CNN with $t={1,2,3...n}$ then,
\begin{equation}
\label{eqn:layer}
        \boldsymbol{\phi_c^t} =\mathbf{ H(I_c^t)}~~,~~ 
        \mathbf{f_c^t} = S(\mathbf{F_c^t})
\end{equation}
If features from layer $l$ are expressed as $\psi$, then 
\begin{equation}
\label{eqn:intermediate}
\boldsymbol{\psi_c^t} = H_l(\mathbf{I_c^t}) 
\end{equation}
 The features from Eq.~\ref{eqn:intermediate} are then pooled in the channel dimension to produce $N \times 1 \times I \times J$  feature for attention in the spatial dimension. The features are then activated by an activation function to produce a spatial soft attention map($A(f_c^t)$).
where $A$ is the sigmoid activation function, and $a_c^t$ is the output of the activation function. Finally attention is applied to the intermediate features $\psi$ at an intermediate layer to obtain activated features $\Psi$ by element-wise multiplication with the activation $a_c^t$.
\begin{equation}
\label{eqn:final}
    \boldsymbol{\Psi_c^t} = \boldsymbol{\psi_c^t} \odot \mathbf{ a_c^t}
\end{equation}  

After gated attention of features at the intermediate CNN layers, the feature extraction process is continued with the rest of the CNN layers. In Eq.~\ref{eqn:last_feat} ${l,k}$ represents layers between intermediate layer $l$ and last layer $k$. Then, the output $\phi_c^t$ of feature extraction is given by,
\begin{equation}
\label{eqn:last_feat}
     \boldsymbol{\phi_c^t} = H_{l,k}(\boldsymbol{\Psi_c^t})
\end{equation}
$\phi_c^t$ is further average pooled in temporal dimension to produce CNN features for re-identification. The network is trained with classification layer similar to our Mutual Attention Network. The Gated Attention method" described here served as one of the baselines for flow guided attention.

%%%%%%%%%%%%%
\section{Results and Discussion}
We start this section with some ablation study on contribution of different modules for the final result. We also include some empirical study on selecting the Deep Feature Extractor layer for for attention as well as sequence length selection. We then proceed to overall performance comparison with the baselines and finally a comparison with state-of-the-art-methods. We do the ablation study on baseline ResNet50 architecture based system~\cite{revisiting} with temporal pooling for feature aggregation on MARS dataset. We compare the overall performance on both MARS and Duke-MTMC dataset.

\begin{table}[b!]

\caption{Accuracy of our Baseline (ResNet50 + Temporal Pooling, TP) and ou Baseline + Mutual Attention(MA) + TP (Ours) at different layers of ResNet50 on MARS dataset.}
\begin{center}
\scalebox{1}
{
\begin{tabular}{|l||l||l|}
\hline
\multicolumn{1}{|c||}{\textbf{Method}} & \textbf{maP} & \textbf{\begin{tabular}[c]{@{}l@{}}  Rank-1\end{tabular}} \\ \hline \hline
Baseline~\cite{revisiting}                             & 75.8         & 83.1                                                           \\ \hline
Baseline+MA (Layer2)        & 78.2         & 84.5                                                           \\ \hline
Baseline+MA (Layer3)        & 78.8         & 84.9                                                           \\ \hline
Baseline+MA (Layer4)        & \textbf{80.0}         &  \textbf{86.6}                                                           \\ \hline
Baseline+MA (Layer5)        & 78.1         & 84.3                                                           \\ \hline

\end{tabular}
}
\end{center}

\label{tab:layer}
\vspace{-0.3cm}
\end{table}
 
\begin{table}[b]
\caption{An ablation study of contribution of different module i.e Gated Attention(ours) and our Mutual Attention on the baselines. This study was done on  MARS dataset}
\vspace{-0.3cm}
\begin{center}
\scalebox{0.99}
{
\begin{tabular}{|l||l||c|c|}
\hline
\textbf{Method} & \textbf{\begin{tabular}[c]{@{}l@{}}Feat \\ Aggregation\end{tabular}} & \multicolumn{1}{l|}{\textbf{mAP}} & \multicolumn{1}{l|}{\textbf{Rank 1}} \\ \hline
\begin{tabular}[c]{@{}l@{}}Baseline~\cite{revisiting} \\ (One Stream)\end{tabular} & Pooling & 75.8 & 83.1 \\ \hline

\begin{tabular}[c]{@{}l@{}}No Attention \\ (Two Stream)\end{tabular} & Pooling & 76.7 & 84.3 \\ \hline  
\begin{tabular}[c]{@{}l@{}}Gated Attention \\ (One Stream)\end{tabular} & Pooling & 77.4 & 84.6 \\ \hline
\begin{tabular}[c]{@{}l@{}}Mutual Attention \\ (Two Stream)\end{tabular} & Pooling & 79.1 & 85.4 \\ \hline  \hline

Baseline~\cite{revisiting} & RNN & 75.7 & 82.9 \\ \hline
Baseline~\cite{revisiting} & \begin{tabular}[c]{@{}l@{}}Temporal \\ Attention\end{tabular} & 76.7 & 83.3 \\ \hline

\begin{tabular}[c]{@{}l@{}}Mutual Attention \\ (Mutual Atten)\end{tabular} & \begin{tabular}[c]{@{}l@{}}Weighted\\  Addition\end{tabular} & \multicolumn{1}{l|}{~\textbf{80.0}} & \multicolumn{1}{l|}{~~~\textbf{86.6}} \\ \hline
\end{tabular}
}
\end{center}

\label{tab:ablation} 
\vspace{-.5cm}

\end{table}

%\subsection{Ablation Study}
%\vspace*{-3mm}
%\paragraph{a) Flow Guided Attention Fusion.}
\noindent{\textbf{a) Flow Guided Attention Fusion.} }
The first part of our work consists of flow guided attention on the intermediate layer of the Deep CNN used for feature extraction in ReID. Different layers in the Deep CNN have different abstraction level of salient features of the input person image. Hence an experiment was conducted by fusing the flow guided attention at different layers of the Depp CNN applied on the baseline~\cite{revisiting} ReID system and evaluated on MARS dataset. From the Tab.~\ref{tab:layer} we conclude that the best performance was achieved by attending at layer 4 of the ResNet50 network. This is justifiable from the fact that the earlier layers have different abstraction level of the salient features, the abstraction level increases in the deeper layers but in the last layer the spatial coherence is lower than the previous layers.

%\vspace*{-3mm}
%\paragraph{b) Contribution of Different Modules to the Baseline.}
\noindent{\textbf{b) Contribution of Different Modules to the Baseline.}} 
In this subsection we compare  Mutual Attention methods to the baseline~\cite{revisiting}  since we follow similar overall system architecture. In Tab.~\ref{tab:ablation} we compare the baseline and ours with different feature aggregation methods like Average Pooling, Temporal Attention and our weighted feature addition method described earlier. We also compare our Mutual Attention method with single stream with Gated Attention using optical flow described in the introduction to Experiments section. We can see that just Gated Attention on its own on the baseline~\cite{revisiting} has improved the performances by a large margin  on MARS datasets. Our Mutual Attention method further improved the results compared to Gated Attention showing the potential of Mutual Attention between both image and optical flow features. Out Feature addition method used with Mutual Attention improve the results for feature aggregation by a larger margin compared to both Average Pooling and aggregation method from Gated Attention. 

\begin{table}[]
\vspace{-0.1cm}
\caption{Empirical study of video sequence length vs performance of our  proposed Mutual Attention (MA),  and baselines with no attention, average pooling and with RNN on MARS dataset.}
\vspace{-0.23cm}
\begin{center}
\scalebox{0.62}
{
\begin{tabular}{|r||c|c|c|c|c|c||c|c|}
\hline
\multicolumn{1}{|l|}{\multirow{3}{*}{\textbf{\begin{tabular}[c]{@{}l@{}l@{}}No of \\ frames \\ per \\Seqence\end{tabular}}}} & \multicolumn{2}{c|}{\textbf{Baseline~\cite{revisiting}}} & \multicolumn{2}{c|}{\multirow{2}{*}{\textbf{\begin{tabular}[c]{@{}c@{}}RNN\\ Aggreg\end{tabular}}}} & \multicolumn{2}{c|}{\textbf{Ours}} & \multicolumn{2}{c|}{\textbf{Ours}} \\ \cline{2-3} \cline{6-9} 
\multicolumn{1}{|l|}{} & \multicolumn{2}{l|}{\textbf{\begin{tabular}[c]{@{}l@{}}Average Pooling\\  No Attention\end{tabular}}} & \multicolumn{2}{c|}{} & \multicolumn{2}{l|}{\textbf{\begin{tabular}[c]{@{}l@{}}Gated Attention\\ Weighted \\ Addition.\end{tabular}}} & \multicolumn{2}{l|}{\textbf{\begin{tabular}[c]{@{}l@{}}Mutual Attention\\ Weighted \\ Addition.\end{tabular}}} \\ \cline{2-9} 
\multicolumn{1}{|l|}{} & \multicolumn{1}{l|}{MAP} & \multicolumn{1}{l|}{Rank 1} & \multicolumn{1}{l|}{mAP} & \multicolumn{1}{l|}{Rank 1} & \multicolumn{1}{l|}{mAP} & \multicolumn{1}{l|}{Rank 1} & \multicolumn{1}{l|}{mAP} & \multicolumn{1}{l|}{Rank 1} \\ \hline \hline
2 & 71.0 & 81.8 & - & - & 77.0 & 84.0 & 74.8 & 82.4 \\ \hline
4 & 75.1 & 83.2 & 75.7 & 82.9 & 77.8 & 84.8 & 77.7 & 85.4 \\ \hline
6 & 74.4 & 82.7 & - & - & 77.6 & 84.5 & 79.2 & 85.8 \\ \hline
8 & 73.3 & 82.0 & 76.2 & 82.5 & 77.3 & 84.2 & 79.3 & 86.4 \\ \hline
16 & - & - &  &  & 72.5 & 82.9 & \textbf{80.0} & \textbf{86.6} \\ \hline
\end{tabular}
}
\end{center}
\vspace{-.6cm}
\label{tab:seqlen} 

\end{table}

%\vspace*{-3mm}
%\paragraph{c) Effect of Sequence Length.}
\noindent{\textbf{c) Effect of Sequence Length.}} 
The length of the sequence has an effect on the representative power of final aggregated feature. This in-turn influences the performance of various feature aggregation methods. Therefore in this subsection we analyse the effect of sequence length on different feature aggregation methods such as Temporal Pooling, Flow guided weighted Addition applied on our method. Hence in Tab.~\ref{tab:seqlen} we have shown results of flow guided attention on ResNet50 architecture with both the feature aggregation methods. It can be seen that at the sequence length of 4 we obtain ideal results for most methods in the literature as well as for the simple gated attention method. But our Mutual Attention method demonstrate the ability to aggregate additional features and hence we could use a sequence of length 16 with Mutual Attention. This is a crucial result as we demonstrate ability to aggregate additional features and keep improving results until a sequence length of 16. Longer sequences have attributed to long term better motion and appearance features. At the same time in other methods in literature, simple averaging adds additional noise to the features with longer sequences. Our weighted addition methods weights the individual feature based on importance and relevance thereby reducing noise with longer sequences. 

\begin{table}[t!]

\caption{Accuracy of our proposed method vs state-of-the-art evaluated on MARS and Duke-MTMC dataset.* next to the method indicates use of optical flow as one of the streams of input.}
\vspace{-0.2cm}

 %From the results it can be concluded that our weighted feature addition  method has out performed Average Pooling and temporal Attention since we have the advantage of both flow features and image features while calculating weights for aggregation.
\begin{center}
\scalebox{0.67}
{
\begin{tabular}{|l|l|c|c||c|c|}

\hline
{\textbf{Method}} &
  {\textbf{Reference}} &
   
  \multicolumn{2}{c|}{\textbf{MARS}} &
  \multicolumn{2}{|c|}{\textbf{Duke-MTMC}} \\ \cline{3-6} 
 
   &
  \multicolumn{1}{l|}{} &
  \multicolumn{1}{l|}{mAP} &
  \multicolumn{1}{l|}{Rank-1} &
  \multicolumn{1}{l|}{mAP} &
  \multicolumn{1}{l|}{Rank-1} \\ \hline\hline
LOMO+SQDA~\cite{liao2015person}           & CVPR-2015  & 16.4  & 30.7 & -     & -     \\ \hline
Set2set~\cite{set2set}           & CVPR-2017  & 51.7  & 73.7 & -     & -     \\ \hline
JST RNN~\cite{joint-spatial}           & CVPR-2017                      & 50.7  & 70.6 & -     & -     \\ \hline
k-reciprocal~\cite{zhong2017re}      & CVPR-2017                        & 58.0    & 67.8 & -     & -     \\ \hline
TriNet~\cite{hermans2017defense}            & ArXiv                             & 67.7  & 79.8 & -     & -     \\ \hline
RQEN~\cite{regionbased}              & AAAI-2018                    & 71.14    & \textbf{77.83} & -     & -     \\ \hline \hline
Part-Alignment~\cite{part_alighned}    & ECCV-2018                       & 72.2  & 83.0   & 78.34 & 83.62 \\ \hline
Mask-Guided~\cite{maskguided}       & CVPR-2018                       & 71.1 & 77.1 & -     & -     \\ \hline
Snippet*~\cite{snippet}            & CVPR-2018                  & 71.1  & 82.1 & -     & -     \\ \hline

STA~\cite{sta}            & AAAI-2019                    & 80.8  & 86.3 &  94.9 & 96.2     \\ \hline

RevstTemPool~\cite{revisiting}      & Arxv       & 75.8  & 83.1 & -  & -  \\ \hline
Cosam-ResNet50~\cite{coseg}    & ICCV-2019                       & 76.9  & 83.6 & 93.5  & 93.7  \\ \hline
STAL~\cite{STAL}    & IEEE Transaction                         & 73.5  & 82.2 & -  & -  \\ \hline
Cosam-SE-ResNet50~\cite{coseg} & ICCV-2019                     & 79.9  & 84.9 & 94.1  & 95.4  \\ \hline 
STAR*~\cite{star}              & BMVC-2019                    & 76.0    & 85.4 & -     & -     \\ \hline \hline
SCAN~\cite{scan}              & IEEE Transaction                    & 76.7    & 86.6 & -     & -     \\ \hline \hline
SCAN~\cite{scan}              & IEEE Transaction                    & 77.2    & \textbf{87.2} & -     & -     \\ \hline \hline
Rec3D~\cite{rec3d}            & IEEE Transaction                   & 80.4  &  86.3& -     & -     \\ \hline
GLTR~\cite{gltr}            & CVPR 2019                    & 78.47  & 87.02 & 93.74   & 96.29     \\ \hline
\begin{tabular}[c]{@{}l@{}}Mutual Attention(Ours) \\ResNet-50 \end{tabular}   &   \textbf{Ours}                           & \textbf{80.0}  & 86.6 & 94.9  & 95.6  \\ \hline  
\begin{tabular}[c]{@{}l@{}}Mutual Attention(Ours) \\SE-ResNet-50 \end{tabular}   &   \textbf{Ours}                          & \textbf{80.9}  & \textbf{87.3} & 94.8  & \textbf{96.7}  \\ \hline  

\end{tabular}
}
\end{center}

\label{tab:soa}
\vspace{-.4cm}
\end{table}

\begin{table}[]
\caption{Comparison of the performance of our work with state-of-the-art methods evaluated on ILIDS-Vid dataset.}
\vspace{-0.1cm}
\begin{center}
\scalebox{1}
{
\begin{tabular}{|l||l||c|c|}
\hline
\textbf{Method} & \textbf{\begin{tabular}[c]{@{}l@{}}Reference \end{tabular}} &  \multicolumn{1}{l|}{\textbf{Rank 1}}&  \multicolumn{1}{l|}{\textbf{Rank 5}} \\ \hline

\begin{tabular}[c]{@{}l@{}}Two \\ Stream*~\cite{optical1} \end{tabular} & ICCV 2017 & 60.0 & 86.0 \\ \hline

\begin{tabular}[c]{@{}l@{}}Snippet~\cite{snippet} \end{tabular} & CVPR-2018 & 85.4 & 87.8 \\ \hline
\begin{tabular}[c]{@{}l@{}}RQEN~\cite{rquen} \end{tabular} & AAAI-2018 &  77.1  & 93.2\\ \hline
\begin{tabular}[c]{@{}l@{}}STAL~\cite{rquen} \end{tabular} & IEEE Trans &  82.8 &95.3 \\ \hline
\begin{tabular}[c]{@{}l@{}} COSAM\\SE-ResNet50~\cite{coseg}  \end{tabular}  & ICCV-2019                        & 79.6 & 95.3     \\ \hline
\begin{tabular}[c]{@{}l@{}}GLTR~\cite{gltr} \end{tabular} & CVPR-2019 &   86.0 & -\\ \hline
\begin{tabular}[c]{@{}l@{}}Rec3D~\cite{rec3d} \end{tabular} & IEEE Trans &   87.9 &98.6 \\ \hline
\begin{tabular}[c]{@{}l@{}}Mutual Attention \\ ResNet-50 \end{tabular} & Ours  &  86.2 & 96.4 \\ \hline
\begin{tabular}[c]{@{}l@{}}Mutual Attention \\ SE ResNet-50 \end{tabular} & Ours  & \textbf{ 88.1} & 98.4  \\ \hline

\end{tabular}
}
\end{center}

\label{tab:ilids} 
\vspace{-.5cm}

\end{table}

\noindent{\textbf{d) Comparison with State-of-the-Art.}}
We  report the performance of our method with backbones ResNet50 and SE-ResNet50~\cite{seres} separately with our Mutual Attention and weighted feature aggregation method on MARS , Duke-MTMC datasets and ILIDS-VID~\cite{ilids} in the Tab.~\ref{tab:soa} and Tab.~\ref{tab:ilids} compared with related works.  As mentioned earlier we have selected~\cite{revisiting} as our baseline. It can be observed that from our baseline. we have improved by a large margin on both mAP and Rank1 metric. Our method has also outperformed most of the state-f-the-art methods including some of the best existing methods. We have also shown the advantage of our method compared to other optical flow based methods~\cite{scan, star,snippet}. Although ~\cite{nlb} have demonstrated State-Of-the-Art results, we do not compare with them as their evaluation strategy is different from that of the commonly followed method in literature. We attribute our performance gain compared to the baseline on both flow guided attention and our feature aggregation technique.
It can also be observed that from our proposal Mutual Attention method performs best demonstrating that optical flow and image stream can attend to the salient regions of each other. Our proposed Mutual Attention method could be integrated with any back-end architecture from some of the strong baselines to achieve better performance. Since our main aim was to evaluate the contribution of Mutual Attention clearly, we choose a simple baseline ~\cite{revisiting}.
Compared to the results on other datasets, the margin of improvement is higher with ILIDS-VID dataset due uniformly centered hand-annotated person cutouts of ILIDS-VID dataset. Results also suggest that our proposed approach is vulnerable to quality of person detection and tracking.

\begin{figure}[]
 \centering
\includegraphics[width=1.0\linewidth]{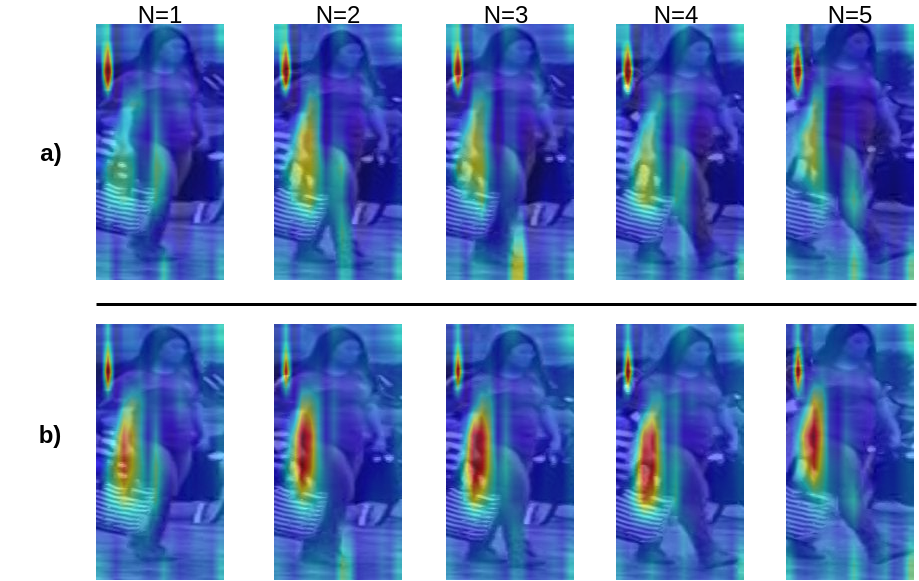}
   \caption{Visualization of feature maps produced using a 5-image video clip selected from the MARS dataset. The video clip on the top row (a) shows the activations of the feature maps without any attention, while the bottom row shows (b) the activations of feature maps with flow guided attention.}
  \label{fig:visuals} 
   \vspace{-0.55cm}
\end{figure}

\noindent{\textbf{d) Visualization.}} Fig.~\ref{fig:visuals} shows activations of feature maps from final layer of our backbone 2D-CNN feature extractor. The top row (a) shows activations without any attention, and the bottom row (b) shows the activations after flow guided attention. It can be observed that our flow guided attention has enhanced the spatial activations of feature maps based on the optical flow produced by the virtue of motion of the person between different frame of the sequence.

%%%%%%%%%%%%%%
\section{Conclusion}
In this work we present a novel framework for flow guided attention and temporal feature aggregation for Person-ReID. The sole purpose of the work has been to focus on visual appearances across spatial and temporal streams and their correlations to encode common saliency between the streams, reduce background clutter, learn motion patterns of person and to have the advantages of having longer sequences. Our feature aggregation method uses cues from both image and optical flow feature to assign weights and aggregate image instance features to produce a single video feature representation unlike assigning equal weights to images instances as in temporal pooling. Our method outperforms the state-of-the-art person reID methods in terms of both mAP and Rank1 accuracy evaluated on MARS, Duke-MTMC, and ILIDS-VID datasets. The proposed Mutual Attention network is most effective when the person detection and tracking produces high quality bounding-boxes, and in scenarios with bigger-sized bounding boxes for objects, where the attention helps in locating the objects. 

\FloatBarrier

{\small
\bibliographystyle{ieee_fullname}
\bibliography{egbib}
}

\end{document}